\documentclass{template}
\usepackage[ruled,vlined]{algorithm2e}
\usepackage{multicol}
\usepackage{pdfpages}
\usepackage[american]{babel}
\usepackage[mathscr]{euscript}
\usepackage{graphicx}
\usepackage{bbold}
\usepackage{blindtext, subfig}
\usepackage{dblfloatfix}
\usepackage{mathtools}
\usepackage{amsmath}
\usepackage{float}
\DeclareMathOperator*{\argmax}{arg\,max}

\usepackage{natbib} 
\usepackage{mathtools} 
\usepackage{booktabs} 
\usepackage{tikz} 

\title{LST: Lexicon-Guided Self-Training for Few-Shot Text Classification}

\author[1]{Hazel Kim\dag} 
\author[1]{Jaeman Son\dag}
\author[1]{Yo-Sub Han}

\affil[1]{%
    Yonsei University\\
    Seoul, Korea
}
\affil[$\dag$]{%
    \scriptsize Equal Contribution
}
\begin{document}
\maketitle

\begin{abstract}
Self-training provides an effective means of using an extremely small amount of labeled data to create pseudo-labels for unlabeled data. Many state-of-the-art self-training approaches hinge on different regularization methods to prevent overfitting and improve generalization. Yet they still rely heavily on predictions initially trained with the limited labeled data as pseudo-labels and are likely to put overconfident label belief on erroneous classes depending on the first prediction. To tackle this issue in text classification, we introduce \textbf{LST}, a simple self-training method that uses a lexicon to guide the pseudo-labeling mechanism in a \emph{linguistically-enriched} manner. We consistently refine the lexicon by predicting confidence of the unseen data to teach pseudo-labels better in the training iterations. We demonstrate that this simple yet well-crafted lexical knowledge achieves 1.0-2.0\% better performance on 30 labeled samples per class for five benchmark datasets than the current state-of-the-art approaches.
\end{abstract}

\section{Introduction}\label{sec:intro}

Due to the high acquisition cost of labeled data, semi-supervised learning is in demand for natural language processing tasks. Self-training, one of the earliest semi-supervised methods, has led to significant improvements in text classification when there is a limited number of labeled data. Self-training uses a teacher model trained from the labeled data and creates pseudo-labels for unlabeled data, which are often larger than the labeled data~\citep{scudder1965probability, yarowsky1995unsupervised}; we call this a pseudo-labeling mechanism. 

Recent self-training approaches adopt several intricate methods within the pseudo-labeling mechanism and regularize well on the unseen data: (1) manipulating a loss term computed on labeled or unlabeled data to generalize the model better for unseen data~\citep{berthelot2019mixmatch, laine2016temporal, rasmus2015semi}; (2) re-weighting samples to select potentially more informative and representative unlabeled data~\citep{wang2020adaptive, panagiota2019enhancing, li2019learning, chang2017active}; (3) adding noise while augmenting data to reduce overfitting under high confidence values~\citep{verma2019interpolation, xie2019unsupervised}. 
Despite their empirical success, these techniques are often wieldy and complicated to implement.

We propose \textbf{L}exicon-Guided \textbf{S}elf-\textbf{T}raining~(\textbf{LST}): a simple self-training method that uses a guidance of lexicon to teach the pseudo-labeling mechanism in a \emph{linguistically-enriched} manner. We build the lexicon with a golden labeled data to guide the whole process of learning. While training models, we inject more data augmented with lexical knowledge to the lexicon. We observe that the more refined lexicon outperforms the state-of-the-art approaches employing self-training with various combinations of complex mechanisms.
We target a simpler yet more accurate approach that leads to the following benefits: 

\begin{itemize}
\item We show that the lexicon, in addition to the teacher model confidence, enhances the reliability of pseudo-labels by manipulating a loss term.
\item We demonstrate that the lexicon selectively samples and possesses more informative unlabeled data to influence the teacher model in a way in which it generates favorable pseudo-labels to the student model.
\item We further show that data augmentation using lexical knowledge such as WordNet builds the lexicon powerful to unseen examples on par with initially annotated data. The lexical knowledge includes semantically different levels of word hierarchy: synonymy, hypernymy, hyponymy, and antonymy.
\end{itemize}

We verify our method using five benchmark data consisting of different numbers of classes. The experimental results confirm that LST outperforms state-of-the-art performance in text classification when labeled data is scarce and unlabeled data is abundant.

\section{Related Work}
Our approach incorporates a reliable pseudo-labeling mechanism using a lexicon, 
and improves a few-shot self-training for text classification. 
Self-training is a semi-supervised method based on a teacher-student framework~\citep{xie2020self}. 
The teacher model, trained from labeled data, creates pseudo-labels for unlabeled data. Then we re-train the 
model using both 
the initial labeled data and the pseudo-labeled data. 
The resulting model becomes a student model  that 
would have a better performance than the teacher model.

Self-training achieves significant improvements especially 
when a pseudo-labeling mechanism generalizes 
the model well on unseen data. 
Current high-performance methods 
include 1) confidence regularization on the teacher model to obtain better pseudo-labels, 2) selective sampling on the unlabeled data to determine which examples influence the teacher model first, and 
3) injecting noise to the model to adapt well on the unseen data. 
We review these methods and illustrate how our lexicon would improve previous approaches to achieve better performance.

\subsection{Confidence Regularization} 
The essence of self-training is entropy minimization, leading teacher model outputs to be sharp pseudo-labeled data and student model outputs to be as sharp as labeled data. Yet the initially ideal teacher model does not always generate perfect pseudo-labels, and a precise prediction from the student model does not guarantee impeccable accuracy~\citep{zheng2021rectifying}. 
Accordingly, trusting all selected pseudo-labels as ground truth can lead to overconfident mistakes and propagated errors~\citep{zou2019confidence}. This issue motivates a confidence regulator to prevent infinite entropy minimization only on annotated data the teacher model initially predicts more confident~\citep{berthelot2019mixmatch}. Instead, many researches~\citep{laine2016temporal, rasmus2015semi, kumar2010self} demonstrate that entropy minimization that is computed on unlabeled data, in addition to labeled ones, encourages the model to output confident predictions and generalize well on the unseen data. Another method~\citep{li2019learning, zou2019confidence} is to generate soft pseudo-labels that redistribute a certain amount of confidence to other classes. Even though the confidence on the pseudo-labels is expected to be precise, the soft pseudo-labels attenuate the misleading effect brought by incorrect or ambiguous supervision. Prior researches indeed show that self-training without regularization is unfavorable to seeking incentives for the model to change and outputs the original model as an optimal capability to do~\citep{kumar2020understanding}. 

\subsection{Selective Sampling}

An important ingredient for self-training is the unannotated data~\citep{du2020self}. Self-training builds a model for identifying and utilizing potentially informative and representative unlabeled samples~\citep{panagiota2019enhancing}. Recent approaches use sample selection to generate more confident pseudo-labels, critical to enhance the learning performance. Sample selection uses the easiness of the samples to inform a learning schedule like training on easy concepts first followed by complex ones. Sample selection leverages teacher confidence~\citep{bengio2009curriculum, kumar2010self} or uncertainty estimates~\citep{mukherjee2020uncertainty} to assess the easiness of a sample. Prior researches demonstrate the model performance depends on training on either easy or complex concepts first. Sample re-weighting~\citep{wang2020adaptive} is another approach to achieve the same goal by mitigating error propagation from noisy pseudo-labels.

\subsection{Noise Injection}
The role of noise has shown to be consistency regularization \citep{sohn2020fixmatch}.
Consistency regularization imposes a constraint on a model to make it harder to memorize the training data and generalize better to unseen data. This procedure enforces the model insensitive to the noise by minimizing consistency loss computed on both original and noised data~\citep{berthelot2019mixmatch}. Accordingly in self-training, noise injection to the student aims to make the student learn beyond the teacher's knowledge. While a powerful unnoised, labeled data is necessary for the teacher model to produce high-quality pseudo-labels, those labels with noise lead the student model to have prediction consistency across unseen data~\citep{ zhang2017mixup}. When applied to unlabeled data, noise has an important benefit of enforcing invariances in the decision function on both labeled and unlabeled data~\citep{xie2020self}. Noise such as stochastic depth, dropout, and data augmentation plays an important role in enabling the student model to have a better capacity than the teacher in self-training.

\begin{figure*}[htb]
  \centering
  \includegraphics[width=1.0\linewidth]{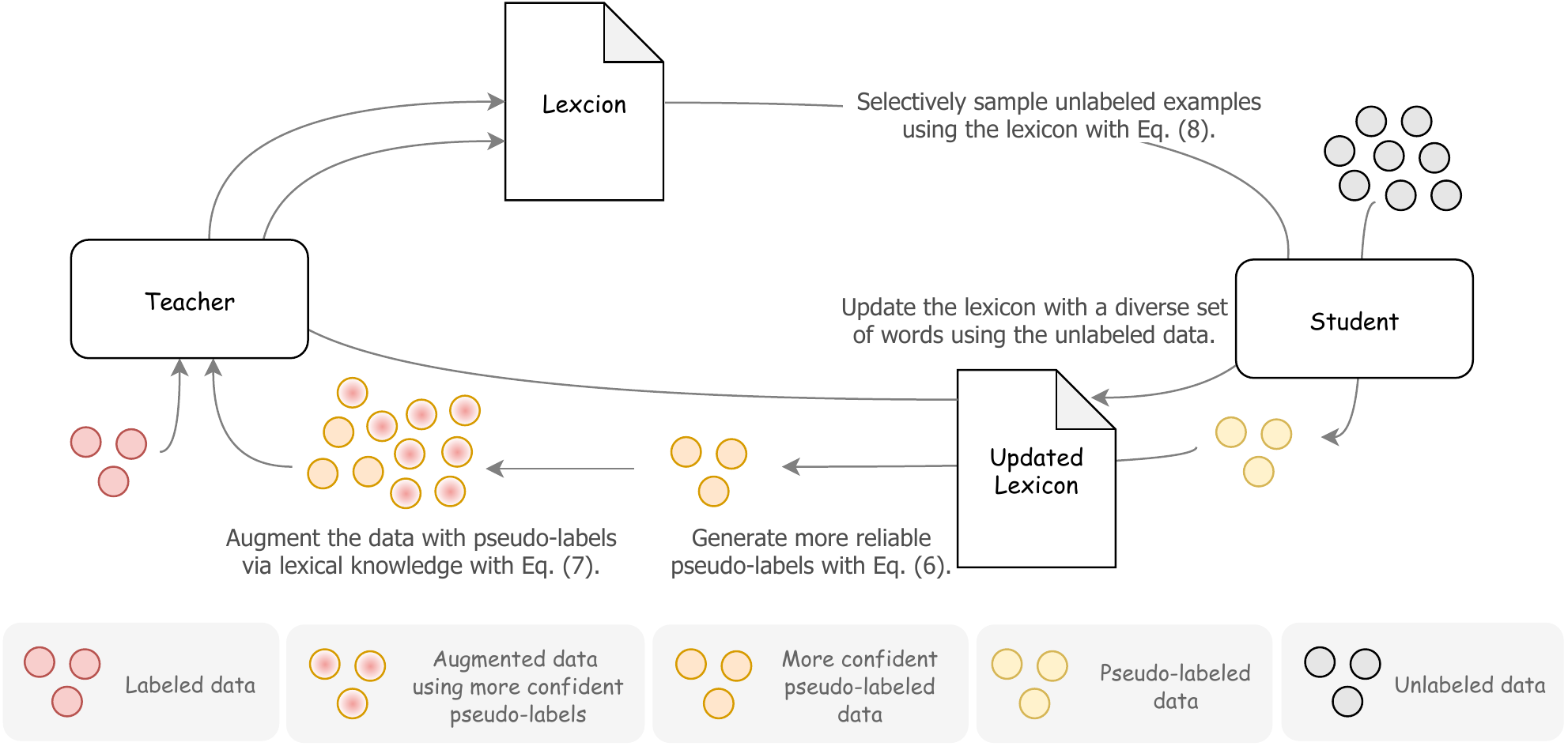}
  \caption{Overview of Lexicon-Guided Self-Training (LST)}\label{fig:lexset_main}
\end{figure*}

Data augmentation creates new data by applying a transformation to an example without changing its label~\citep{xie2020self}---this becomes a helpful way of injecting noise in self-training~\citep{verma2019interpolation, xie2019unsupervised}. Strong augmentation methods generate diverse examples that share the same ground-truth labels with original examples. Such data augmentation methods include bidirectional language model-based replacement by randomly replacing target words~\citep{Kobayashi18} and BERT masked language model-based replacement by predicting new words using the context~\citep{WuLZHH19, NgCG20}. Another approach without model-based is a simple word replacement using synonyms, neighbor words or superior words through large-scale lexical database such as WordNet~\citep{Miller95, wei2019, ZhangZL15, xu_2019}. The diverse set of augmented examples encourages the model to learn consistency on unseen data and improve self-training performance. 
\section{Lexicon-Guided Self-Training}

We extend the traditional self-training to use a lexicon, which is a set of representative words for each class in a specific domain. The lexicon, crafted by frequent occurrence of semantically important words, enhances the reliability of the pseudo-labeling mechanism. 
The lexicon in self-training becomes an additional criteria for selecting data to fulfill qualifications of 
1)~being a pseudo-label, 2)~being the next unlabeled example, and 3)~being a meaningful noise added to the student model. 
With the guidance of lexicon, our approach iteratively executes the following procedures until the model converges to a certain standard of performance outcome. 
  
\subsection{Lexicon Construction}
\label{sec:lex_con}
A lexicon is a set of representative words for a class. 
Our lexicon construction is as follows.  
For each example, we choose the top $n$~words that have the 
highest 
attention scores in a fine-tuned BERT-LSTM classifier 
because these words are closely associated with the corresponding class label. 
Among the selected words, we remove words that appear in 
examples with different classes; in other words, we only keep 
the words with distant classes.
Because high attention scores do not necessary imply 
representative words for text classification,  
we also consider the word frequency.
For each class, we count the number of 
word occurrences in all examples of the  class,
and select the top $k\%$ frequent 
words---the selected words are a lexicon of the class.

\subsection{Pseudo-Labeling Guided by Lexicon}

\begin{algorithm*}
\SetAlgoLined
 Continue pre-training language model on task-specific unlabeled data; \\
 Fine-tune model $f^{w}$ with parameters $\textbf{w}$ on task-specific small labeled data with Equation~\eqref{firstTeacher}; \\
 Construct an initial lexicon using labeled data with Equation~\eqref{firstPseudo}; \\
 \While{not converged}{
  \For{$\textbf{x}_t \in \textbf{X}_T$ }  
     {Follow pseudo-labeling procedure with teacher model confidence $f^{w}$ and lexicon regulator $\mathscr{R}_{lex}$ with Equations~\eqref{PseudoCond},~\eqref{PseudoConst}, and~\eqref{lexReg}; \\
     Pseudo-label on $\textbf{x}_t$ using the model $f^{w}$ with Equation~\eqref{lossCR}; \\
     }
   \For{  $\{\hat{\textbf{y}}_t \cup \textbf{x}_t\} \in T$   } 
   { 
    Inject noise to the lexicon using lexical knowledge for data augmentation; 
   }
   Re-train the model $f^{w}$ with the augmented data and update parameters $w$ with Equation~\eqref{lossDA}; \\
   Extract refined words and update the lexicon; \\
   Make the student model as a teacher with Equation~\eqref{lossSS}; \\
  }
\caption{Lexicon-Guided Self-Training (LST)}
\end{algorithm*}

LST performs a pseudo-label estimation under a unified loss minimization problem using lexicon. 
Pseudo-labels are treated as discrete learnable latent 
variables being either one-hot or all-zero~\citep{zou2019confidence}. Given a target example  $\textbf{x}_t$ from unlabeled target domain data $\textbf{X}_T$, we compute a target label $\hat{\textbf{y}}_t = (\hat{\textbf{y}}_t^{(1)}, \ldots ,\hat{\textbf{y}}_t^{(C)} )$ with respect to labeled data  $( \textbf{x}_s, \textbf{y}_s )$ from source domain  $( \textbf{X}_S, \textbf{Y}_S )$. This step of computing target label $\hat{\textbf{y}}_t$ is for generating pseudo-labels by a teacher model. Then, we make the teacher model as a student. We train the student model with both labeled data $( \textbf{x}_s, \textbf{y}_s )$ and pseudo-labeled data~$(\textbf{x}_t, \hat{\textbf{y}}_t)$, to tune the network weights \textbf{w} using $p(c|\mathbf{x}; \mathbf{w})$, a classifier's softmax probability for class $c$. Here we explain two steps of generating initial pseudo-labels from a teacher model before they go to the loop of iterating the teacher-student learning procedure: (1) teacher model optimization; (2) pseudo-label generation. 


\paragraph{(a) Teacher Model Optimization.} 
As for the first step, we fix 
$\mathbf{Y}_s$, and train a teacher model with labeled data. We optimize the teach model by 
fixing $\mathbf{Y}_s$ and solving:

\begin{equation}
\label{firstTeacher}
   \mathbf{w} = \min \Big( - \sum_{s \in S} \sum_{k=1}^{c} {y_s^{(c)}\log{p(c|\mathbf{x}_s;\mathbf{w})}} \Big).
\end{equation}

\paragraph{(b) Pseudo-Label Generation.} Using the initial teacher model, we generate pseudo-labels for unlabeled data and construct a lexicon. We obtain a lexicon in this procedure following the construction methods explained in Section~\ref{sec:lex_con}. The potential pseudo-labels filtered through the model prediction now consider a lexicon as the next criteria to meet. We can ultimately show the entropy minimization for generating initial pseudo-labels from the teacher model using Equation~\eqref{firstPseudo}. We fix $\mathbf{w}$ and solve:

\begin{equation}
\label{firstPseudo}
\begin{split}
  \hat{\mathbf{Y}}_T & = \min \Big( - \sum_{s\in{S}}\sum_{c=1}^{C}{\mathscr{R}_{lex} {(\hat{y}_t^{(c)})}
  \log\frac{P(c|\mathbf{x}_t;\mathbf{w})}{\lambda_c}} \Big)
  \\
  &\hspace{10mm} \mathrm{s.t.} \hspace{3mm} \hat{y}_t \in{\Delta^{C-1}} \cup{\{\mathbf{0}\}} , \hspace{1mm} \forall{t}. 
\end{split}
\end{equation} 

Note that the feasible set of pseudo-label is either $\{\textbf{0}\}$ or a probability simplex $\Delta^{C-1}$. Each example having a one-hot value of $\hat{y}_t^{(c)}$ as a valid probability can work as a selected pseudo-label, while examples with a value of \textbf{0} for $\hat{y}_t^{(c)}$ cannot. The model minimizes the entropy via selecting pseudo-labels between  $\hat{\textbf{y}}_t^{\dagger{}}$ and \textbf{0} using the model confidence with a parameter for sample selection $\lambda_{c}$ and lexicon regulator $\mathscr{R}_{lex}{(\hat{y}_t^{(c)})}$. The selected pseudo-labels lower the entropy of a single example $t$ by checking $E( \hat{\textbf{y}}_t^{\dagger{}} )$ as shown in Equation~\eqref{PseudoCond}.

\begin{equation}
\label{PseudoCond}
\begin{split}
  &\hat{\textbf{y}}_t^{\ast} = 
  \begin{cases}
    \hat{\textbf{y}}_t^{\dagger{}} & \text{if}\ E( \hat{\textbf{y}}_t^{\dagger{}} ) < E(\textbf{0}); \\
    \textbf{0} & \text{otherwise}.
  \end{cases}
\end{split}
\end{equation} 

The entropy value $E( \hat{\textbf{y}}_t^{\dagger{}} )$ depends on the parameter $\lambda_{c}$ and the lexicon regulator $\mathscr{R}_{lex}{(\hat{y}_t^{(c)})}$ as shown in Euqation~\eqref{PseudoConst}. 
The parameter $\lambda_{c}$ is critical to control the sample selection. The class $c = \argmax_{c^{\ast}}\{\frac{p(c^{\ast}|\textbf{x}_t;\textbf{w})}{\lambda_{c^{\ast}}}\}$ labels each selected example when the example's predication is confident enough. The pseudo-label on the example is valid when the value of $p( c|\textbf{x}_t; \textbf{w})$ is greater than the parameter value $\lambda_c$. The less confident one with the example's predication value less than the parameter $\lambda_c$ cannot be a pseudo-labeled example
and our model does not select it.

\begin{equation}
\label{PseudoConst}
\begin{split}
&E( \hat{\textbf{y}}_t^{\dagger{}} ) =  \min_{\hat{\textbf{Y}}_T} - \sum_{s\in{S}}\sum_{c=1}^{C}\mathscr{R}_{lex}(\hat{y}_t^{(c)})\log\frac{P(c|\textbf{x}_t;\textbf{w})}{\lambda_c}
\\
&\mathrm{s.t.~~} c = \argmax_{c^{\ast}} \frac{p(c^{\ast}|\textbf{x}_t;\textbf{w})}{\lambda_{c^{\ast}}}  \mathrm{~~and~~}  p( c|\textbf{x}_t; \textbf{w}) > \lambda_{c}.
\end{split}
\end{equation} 

In addition to the model confidence, each selected example $\textbf{x}_t$ has to contain at least $\tau$ number of lexicon words. The lexicon confirms the validity of pseudo-labels on those selected examples depending on whether the examples contain sufficient lexicon words as evidences of the label. Since the lexicon consists of words that fulfill the minimum threshold---certain number of occurrences with high attention scores within the source domain $( \textbf{X}_S, \textbf{Y}_S )$, the lexicon regulator $\mathscr{R}_{lex} {(\hat{y}_t^{(c)})}$ works as another qualification for strengthening our pseudo-labeling mechanism. 

\begin{equation}
\label{lexReg}
\begin{split}
  \mathscr{R}_{lex}(\hat{y}_t^{(c)}) = 
  \begin{cases}
    \hat{y}_t^{(c)} & \text{if}\hspace{1mm} \sum_{\text{z}_t \in \textbf{X}_t}\text{count}(\text{z}_t|\text{z}_{lex}) \geq \tau; \\
    \textbf{0} & \text{otherwise}.
  \end{cases}
\end{split}
\end{equation} 

Note that the lexicon regulator only depends on pseudo-labels $\hat{\textbf{y}}_t$. In other words, the lexicon guides the pseudo-labeling mechanism and influences the overall self-training in the following three steps:

\paragraph{Confidence Regularization with Lexicon Words}

The lexicon performs as a regulator $\mathscr{R}_{lex}{(\hat{\textbf{Y}}_T)} $, which determines whether to label on a certain example depending on the informativeness of a pseudo-label~$\hat{\textbf{y}}_t$.
\begin{equation}
\label{lossCR}
\begin{split}
  \min_{\hat{ \textbf{Y} }_{t} }   \mathscr{L}_{CR} &= \mathscr{L}_{tr} {(\textbf{w},\hat{\textbf{Y}}_T)} + \mathscr{R}_{lex}{(\hat{\textbf{Y}}_T)} 
  \\
  & = - \sum_{s \in S} \sum_{c=1}^{c} {y_s^{(c)}\log{p(c|\textbf{x}_s;\textbf{w})}}
  \\ 
  & - \sum_{t \in{T}}  \Big( \sum_{c=1}^{C} \hat{y}_t^{(c)} \log{ p(c|\textbf{x}_s ; \textbf{w} ) } -\mathscr{R}_{lex} {(\hat{\textbf{y}}_t)} \Big).
\end{split}
\end{equation} 
The regulator determines whether a candidate pseudo-label is qualified to be a final pseudo-label at the corresponding epoch. 
This helps the teacher model 
generate more reliable pseudo-labels.

\paragraph{Data Augmentation with Lexical Knowledge}
We compute the student model's predicted class distribution on an augmented version of given pseudo-labels from the teacher model. 
We improve the informativeness of pseudo-labels
via lexical knowledge using 
WordNet denoted by $\alpha (\cdot)$. 

\begin{equation}
\label{lossDA}
\begin{split}
  \min_{\textbf{w} }   \mathscr{L}_{DA} & = \mathscr{L}_{stu} {(\textbf{w}, (\hspace{1mm} { \alpha(\hat{\textbf{Y}}_T))} }
  \\
  & =  - \sum_{s \in S} \sum_{k=1}^{K} {y_s^{(k)}\log{p(k|\textbf{x}_s;\textbf{w})}} \\
  & - \sum_{t \in T} \sum_{k=1}^{K} \alpha({\hat{y}_t^{(k)}) \log{p(k|\alpha(\hat{\textbf{x}_t});\textbf{w})}}.
\end{split}
\end{equation} 

We replace words in an example sentence with relevant words, which preserve the label-compatibility. 
More specifically, the replacement method works over four different actions by exploring the lexical knowledge from WordNet: synonymy, hypernymy, hyponymy, and antonymy. 
Unlike other research, we employ the antonym of antonym to generate more diverse samples by keeping the label compatibility. We augment the data by uniformly executing replacement actions per each example.

With the augmented version of pseudo-labels, the student model trains and produces the updated weight values---this re-retrained model is 
the teacher model
for the next   
pseudo-labeling mechanism in self-training.

\paragraph{Selective Sampling with Lexical Attention} 
The previous iteration provides useful weight values for sample selection since the augmented version of confident pseudo-labels enable the student model to focus on more informative examples. 

Since the student model becomes a teacher model, the updated model is able to 
selectively sample useful unseen data and generate candidate pseudo-labels for the next iteration. 
The lexicon regulator performs just like the previous iteration does 
in the teacher model by  Equation~\eqref{lossCR}. 

\begin{equation}
\label{lossSS}
\begin{split}
  \min_{\hat{ \textbf{Y} }_{t}^{\ast} }   \mathscr{L}_{SS} & = \mathscr{L}_{tr} {(\textbf{w},\alpha (\hat{\textbf{Y}}_T))} + \mathscr{R}_{lex}{(\alpha (\hat{\textbf{Y}}_T))} 
  \\
  & = - \sum_{t \in{T}} \Big( \sum_{k=1}^{K} \alpha (\hat{y}_t^{(k)}) \log{ p(k|\textbf{x}_s ; \textbf{w}_\alpha ) } -\mathscr{R}_{c} {(\alpha (\hat{\textbf{y}}_T))} \Big).
\end{split}
\end{equation} 

This procedure keeps taking the augmented information about pseudo-labels into account while updating the teacher model and refining the lexicon until the model converges. The consistently refined lexical knowledge in the lexicon 
improves the overall self-training mechanism.

\section{Experiments}
We evaluate the proposed approach with respect to 
five benchmark datasets: 
IMDB~\citep{maas2011}, SST-2~\citep{socher2013}, 
Elec~\citep{McAuleyL13}, AG News~\citep{ZhangZL15},
and DBpedia~\citep{ZhangZL15}. 
 
We compare our method and state-of-the-art mothods for data augmentation (SSMBA) and few-shot text classification (UDA, UST).

\begin{table}[h!]
\centering
\caption {Dataset summary (\#W: avg.~word per text)}\label{dataset:stats}
\begin{tabular}{@{}ccccc}
\toprule
Dataset & Class & Train & Test & \#W \\
      \midrule 
SST-2   & 2     & 67K   & 1.8K        & 10     \\
IMDB    & 2    & 25K   & 25K         & 235     \\
Elec    & 2     & 25K   & 25K        & 108     \\
AG News & 4     & 120K  & 7.6K          & 40      \\
DBpedia   & 14  & 560K  & 70K           & 51      \\
 \bottomrule 
\end{tabular}
\end{table}

\begin{table*}[h!] 
  \centering
  \caption{Accuracy comparison of different models for text classification on five benchmark datasets}
  \label{tab:result}
\begin{tabular}{@{}cccccccc@{}}
\toprule
\multicolumn{1}{c}{Dataset} & \multicolumn{1}{c}{All train} & \multicolumn{6}{c}{30 labeled data per class for training and for validation}                                                                                                                        \\ \midrule
\multicolumn{1}{c}{}        & \multicolumn{1}{c}{BERT}      & \multicolumn{1}{c}{BERT} & \multicolumn{1}{c}{Classic ST} & \multicolumn{1}{c}{UDA} & \multicolumn{1}{c}{UST} & \multicolumn{1}{c}{SSMBA+BERT} & \multicolumn{1}{c}{LST (ours)} \\ \midrule
SST-2    & 92.12   & 69.79  & 84.81   & 83.58  & 87.69     & 66.70   & 87.64 (1.63)                               \\ \midrule
IMDB     & 91.7  & 73.03  & 78.97   & 89.30  & 89.21  & 76.10    & \textbf{89.87 (1.05)}                          \\ \midrule
Elec     & 93.46    & 82.92     & 89.92   & 89.64    & 91.27  & 78.61   & 90.24 (1.48)                                  \\ \midrule
AG News    & 92.12   & 80.74  & 84.62   & 85.92  & 88.19  & 86.13   & \textbf{89.24 (0.78)}                          \\ \midrule
DBpedia    & 99.26   & 97.77  & 98.39   & 96.88  & 98.57  & 98.64   & \textbf{98.74 (0.23)}                          \\ \midrule
Average    & 93.73   & 80.85  & 87.34   & 89.06  & 91.00  & 81.13     & \textbf{91.14 (1.03)}                          \\ \bottomrule
\end{tabular}
\end{table*}

\subsection{Experiment Setting}

We follow the conventional implementation recommendations for self-training (ST) methods for fair ST evaluations. In particular, this means we use the same model and training algorithm in the same codebase for all experiments regarding ST. We use the same pre-trained language model BERT as our base encoder or classifier to compare Classic ST, UDA~\citep{xie2019unsupervised}, UST~\citep{mukherjee2020uncertainty}, and SSMBA~\citep{NgCG20} with LST

We use the BERT-based-uncased model (12-layer, 768-hidden, 12-heads, 110M parameters) followed by LSTM (768-hidden) with attention layer. 
We predict the class label by using the attention weights for each word and exploring the context vector that captures the entire input sequence to selectively focus on semantically meaningful words. For this, we apply ReLU (0.1) activation and dropout (0.3) to the context vector and then feed it into linear layer. We use the AdamW optimizer with a learning rate 1e-5 and a weight decay of 0.01. We set $n =  3$ and $k = 10$ specified in the lexicon construction in Section~\ref{sec:lex_con}. We set the confidence threshold $\lambda_k$ 
as 0.8 for all benchmark datasets and a matching threshold~$\tau$ as 2 for all but SST-2, 
in which we set as 1. This is because most data in SST-2 are very short compared with the other benchmark 
datasets and, thus, it is more appropriate to choose less number of words than the others.  
We train the model with a batch size of 16 for 30 epochs and apply the early stopping via validation loss per each epoch. We measure the model performance at the point when we have the highest validation accuracy.

Unless otherwise noted, all our experiments use the same number of limited labeled data: we randomly select 30 examples per class from the unlabeled pool for training and development, respectively.
For each benchmark dataset, we repeat the evaluation three times and report the average.

\subsection{Baselines}

\paragraph{BERT} is one of the most popular pre-trained language models and shows the
state-of-the-art performance for 
various natural language processing 
tasks~\citep{devlin2019}. 
We adopt BERT as our base encoder or teacher model, hence compare it with LST as a baseline.

\paragraph{Classic Self-Training (ST)} starts with a base teacher model trained on the labeled data, and the pseudo-labeled data trains a student model~\citep{yarowsky1995unsupervised}. 
The student-teacher training procedure repeats till the model converges. Our baseline copies student model parameters to use as the teacher model during ST.

\paragraph{UDA} leverages back-translation
for data augmentation in ST. UDA follows similar
principles as virtual adversarial training~\citep{MiyatoMKI17} and consistency training~\citep{LaineA17, SajjadiJT16a} such that the model considers the noise injection to improve the performance. Unlike other baselines, this approach requires auxiliary resources using neural machine translation system to generate the back-translation.

\paragraph{UST} adopts Bayesian active learning by disagreement~\citep{Houlsby:11} using stochastic dropouts to selectively sample unlabeled examples that maximize the information gain between predictions and the model posterior. UST~\citep{mukherjee2020uncertainty} minimizes the model variance not only by focusing more on unlabeled examples difficult to predict but by measuring the predictive variance to selectively focus on pseudo-labeled examples that the teacher is more confident on.

\paragraph{SSMBA} is a self-supervised manifold based data augmentation~\citep{NgCG20}, generating pseudo-labels by using denoising auto-encoders (DAE) \citep{BengioYAV13}. SSMBA uses masked language models~\citep{devlin2019, abs-1907-11692} as DAEs to sample meaningful examples from the underlying natural language distribution and to corrupt and reconstruct the examples as an augmented input. 
\citet{NgCG20} used SSMBA+CNN for evaluating the data augmentation performance. 
In our empirical study, SSMBA+BERT shows better performance than 
SSMBA+CNN when there is few labeled data. Hence, for fair comparison, we choose SSMBA+BERT for baseline.

\subsection{Results and Analysis}

Table~\ref{tab:result} shows that LST sets the new state-of-the-art for few-shot text classification. We observe that UST and LST all perform reasonably well. However, UST requires another dataset to learn more information about the unlabeled pool whereas LST follows a more traditional few-shot ST setting---few labeled data and abundant unlabeled data. In other words, LST guarantees a similar or better performance with less amount of data.

LST substantially outperforms UDA and SSMBA on all of the benchmark datasets. The lexicon constructed for LST performs better than the auxiliary resource of back translation for UDA in terms of data augmentation for few-shot text classification. 
Even though SSMBA has a 
robust data augmentation method for the out-of-domaion problem, 
LST works better than SSMBA when
there is extremely limited labeled data. 
This confirms that our approach is a better solution for the low-resource setting.

We use a fully supervised BERT baseline to measure the highest accuracy we could obtain with abundant labeled data. We also train a BERT baseline only with 30 labeled data to compare with other approaches employing ST mechanism trained on the same number of labeled data to test the performance of few-shot learning. On all of benchmark datasets, LST outperfoms BERT and ST with the limited labeled data. LST outputs pretty competitive accuracy compared with all trained BERT with fully supervised approach. 

LST outputs the best performance on datasets consisted of multiple classes for topic classifications such as AG News or DBpedia. 
For the binary classification, LST shows very competitive accuracy on Elec and SST-2, and the best accuracy on IMDB. Note that IMDB has the largest
number of words per text among the benchmark sets.
This confirms that LST performs well when there are diverse and sufficient number of words, which then help to build an effective lexicon for 
more accurate text classification. 

\paragraph{Word Selections for Lexicon Refinement}

\begin{table}[htb]
\centering
\caption{Different frequency ratio~$k\%$ for the lexicon}
\label{ta:freq} 
\begin{tabular}{@{}lrrrr@{}}
\toprule
         &  1\% & 10\%   &  50\% & 100\%  \\ \midrule
 SST-2   & 86.37                    & 87.64  & 84.45                     & 84.68  \\
 IMDB    & 87.64                    & 89.87  & 88.34                     & 84.37  \\
 Elec    & 88.08                    & 90.24  & 87.15                     & 86.42  \\
 AG News & 88.12                    & 89.24  & 87.88                     & 87.43  \\
DBpedia                      & 98.47                    & 98.74  & 98.55                     & 98.34  \\ \midrule
Average                      & 89.73                   & 91.14 & 89.27                    & 88.24 \\ \bottomrule
\end{tabular}
\end{table}

In general, frequent words with high attention scores are representative words for its data and the corresponding class. Yet it might not be desirable to use all frequent words for lexicon. Thus, we 
study the problem of choosing a right amount of frequent words for the lexicon refinement for better
performance. 
Once we collect top three words from each text that have the 
highest attention score,
we sort these words by their occurrences
and select only the top~$k\%$ words for lexicon. 
In our experiment summarized in Table~\ref{ta:freq}, 
we set $k = 1, 10, 50, 100$ and compute their accuracy.

For $k =  1$, most words in the refined lexicon are highly confident words, yet a relatively small size of  
lexicon have a smaller coverage on unlabeled data and do 
not utilize unlabeled data enough.
We achieve the best performance when $k = 10$; here we have enough confident words and a lexicon of 
reasonable size. 
On the other hand, as $k$ increases, the performance decreases. 
This is because we become to have less confident words as $k$ increases, the lexicon also becomes less accurate. Then, there are more noisy pseudo-labels, which produce more accumulated errors.

The lexicon refinement with frequent words often prevents the lexicon falling into a trap of being 
corrupted because words with high attention scores are not necessarily the representative words for the corresponding classes. Also, it is crucial to select not all but a proper fraction of words (e.g., 10\%)
to avoid the problem of adding noisy pseudo-labels too much.

\begin{table*}[htb]
\centering
\caption{Accuracy comparison of different combinations of LST modules}
\label{table:ablation2}
\begin{tabular}{ccccccc}
\toprule
  Combination            & \multicolumn{1}{l}{SST-2} & \multicolumn{1}{l}{IMDB} & \multicolumn{1}{l}{Elec} & \multicolumn{1}{l}{AG News} & \multicolumn{1}{l}{DBpedia} & \multicolumn{1}{l}{Average} \\ \midrule
 BERT (LM)           & 69.79                                          & 73.03                    & 82.92                                         & 80.74                       & 97.77                       & 80.85                       \\
ST + LM  & 84.81                                          & 78.97                    & 89.92                                         & 84.62                       & 98.39                       & 87.34                       \\ 
ST + LM + lexicon  & 87.04                                          & 88.51                    & 90.36                                         & 88.26                       & 98.28                       & 90.49                       \\
ST + LM + lexicon + aug.~(= LST)                           & 87.64                                          & 89.87                    & 90.76                                         & 89.24                       & 98.74                       & 91.25                       \\ \bottomrule
\end{tabular}
\end{table*}

\subsection{Ablation Study}

Since LST leverages the lexicon for ST, we perform an ablation study to understand how lexicon helps LST obtain state-of-the-art results. We focus on comparing LST with and without data augmentation to highlight the impact of lexicon itself in ST and to confirm the usefulness of data augmentation using lexical knowledge. 

Table~\ref{table:ablation2} shows that LST without data augmentation still outperforms BERT and ST on all the datasets. LST without lexicon improves the average accuracy by $3\%$ compared to ST. The performance improvement confirms that the  refined lexicon produces a more reliable pseudo-labeling mechanism and improves the accuracy. 
Note the accuracy gains obtained from the lexicon is all higher than that of UDA, a data augmentation method for consistent learning, by more than $3\%$. We believe that the lexicon already plays a critical role in guiding the ST as a confidence regulator for the model confidence and as a selective sampler for the unseen data. 

Moreover, the lexical knowledge increases the performance of lexicon approximately by $1\%$ 
by augmenting more informative data.
Due to this data, the student model and lexicon become to have better  ability to identify meaningful data from the unseen data pool.
The augmentation consistently provides more accurate pseudo-labeling mechanism while iterating the learning procedure.

\section{Conclusions}

We have introduced LST, a self-training method that utilizes a lexicon to guide few-shot text classification tasks in a linguistically-enriched manner. LST has enhanced the traditional self-training with the following components and obtained the state-of-the-art performance:

\begin{enumerate}

\item Confidence regularization. The refined lexicon prevents relying on the model confidence only, and guides as an additional criteria 
to be confident on new
pseudo-labels candidates.
Because the lexicon contains words learned from the annotated data, it works as a guidance to improve self-training performance.

\item  Selective sampling on unlabeled data. The lexicon produces high confidence values on more informative pseudo-labels and optimizes models to gain a better learning ability on meaningful unlabeled data.

\item Noise injection using lexical knowledge. The lexicon becomes powerful to guide the learning mechanism on the unseen data by enriching the lexical knowledge in lexicon.

\end{enumerate}

On the whole, the simple but performant lexicon significantly improves self-training mechanism in few-shot text classification.

\bibliography{template}

\end{document}